\documentclass[conference]{IEEEtran}

\usepackage{amsmath,amssymb}
\usepackage{graphicx}
\usepackage{listings}
\usepackage{xspace}
\usepackage[table]{xcolor}
\usepackage{algorithm}
\usepackage{algpseudocode}
\usepackage{todonotes}
\usepackage{mathtools} 
\usepackage{tikz}
\usepackage{hyperref} 
\usepackage{url}
\usepackage{multirow}
\usepackage{ctable}

\definecolor{dkgreen}{RGB}{0,64,0}
\definecolor{ltgray}{RGB}{245,245,245}
\definecolor{mauve}{RGB}{139,0,139}

\newcommand{\tweakedsim}{\raise.17ex\hbox{$\scriptstyle\mathtt{\sim}$}}

\newcommand\summit{Summit\xspace}
\newcommand\axonn{AxoNN\xspace}

\newtheorem{theorem}{Theorem}[section]

\newtheorem{lemma}[theorem]{Lemma}

\begin{document}

\title{AxoNN: An asynchronous, message-driven parallel framework for extreme-scale deep learning}

\author{\IEEEauthorblockN{Siddharth Singh, Abhinav Bhatele}
\IEEEauthorblockA{~\\Department of Computer Science, University of Maryland, College Park, Maryland 20742 USA\\
E-mail: ssingh37@umd.edu, bhatele@cs.umd.edu}
}

\maketitle

\begin{abstract}
In the last few years, the memory requirements to train state-of-the-art neural
networks have far exceeded the DRAM capacities of modern hardware accelerators.
This has necessitated the development of efficient algorithms to train these
neural networks in parallel on large-scale GPU-based clusters. Since
computation is relatively inexpensive on modern GPUs, designing and
implementing extremely efficient communication in these parallel training
algorithms is critical for extracting the maximum performance. This paper
presents \axonn, a parallel deep learning framework that exploits asynchrony
and message-driven execution to schedule neural network operations on each GPU,
thereby reducing GPU idle time and maximizing hardware efficiency.  By using
the CPU memory as a scratch space for offloading data periodically during
training, \axonn is able to reduce GPU memory consumption by four times.  This
allows us to increase the number of parameters per GPU by four times, thus
reducing the amount of communication and increasing performance by over 13\%.
When tested against large transformer models with 12--100 billion parameters on
48--384 NVIDIA Tesla V100 GPUs, \axonn achieves a per-GPU throughput of
49.4--54.78\% of theoretical peak and reduces the training time by 22-37 days
(15--25\% speedup) as compared to the state-of-the-art.

\end{abstract}


\section{Introduction}
\label{sec:intro}
In recent years, the field of deep learning has been moving toward training
extremely large neural networks (NNs) for advancing the state-of-the-art in
areas such as computer vision and natural language processing
(NLP)~\cite{gpt-2,megatronlm,gpt-3}. This surge in popularity of large NNs has
been propelled by the availability of large quantities of data and the
increasing computational capabilities of hardware such as GPUs. However, the
memory requirements to train such networks have far exceeded the DRAM
capacities of modern hardware accelerators. This has necessitated the
development of efficient algorithms to train neural networks in parallel on
large-scale GPU-based clusters.

Computation is relatively inexpensive on modern GPUs and has outpaced the
increase in network bandwidth on GPU-based clusters with size.  Hence, the
design and implementation of efficient communication algorithms is critical to
prevent starvation of GPUs waiting for data to compute on.  Contemporary
frameworks for parallel deep learning suffer from a number of shortcomings in
this regard. Some of them employ bulk synchronous parallel algorithms to divide
the computation of each layer across GPUs~\cite{mesh_tf,megatronlm}. The
synchronization step in these algorithms can be time-consuming as it employs
collective communication on the outputs of each layer, which are fairly large
in size.  Other frameworks try to divide contiguous subsets of layers across
GPUs~\cite{megatronlm-2,zero_3D,narayanan2019pipedream}. Data is then exchanged
between consecutive GPUs using point-to-point communication. In this setting,
such implementations fail to exploit the potential of overlapping communication
and computation due to the use of blocking communication primitives and static
scheduling of layer operations.

In this paper, we present \axonn, a parallel deep learning framework that
exploits asynchrony and message-driven execution to schedule neural network
operations on each GPU, thereby reducing GPU idle time and maximizing hardware
efficiency.  To achieve asynchronous message-driven execution, we implement
\axonn's communication backend using CUDA-aware MPI with GPU-Direct support.
While the general consensus among other frameworks~\cite{megatronlm-2, zero_3D,
narayanan2019pipedream} has been to use NCCL~\cite{nccl} or Gloo~\cite{gloo}
for point-to-point communication, we find that MPI is more suitable for this
task because it offers higher intra-node bandwidth and supports non-blocking
primitives, which are great for asynchrony.  This gives AxoNN an edge over
other frameworks in terms of performance.  To the best of our knowledge, this
is the first work to exploit implement message-driven execution for parallel
deep learning, with prior work using static scheduling schemes due to the
constraints of synchronous communication.

Neural networks with extremely large number of parameters like the
GPT-3~\cite{gpt-3} architecture require a correspondingly large number of GPUs
often ranging in the hundreds. At such large GPU counts, there is a notable
drop in hardware efficiency due to increasing communication to computation
ratios~\cite{megatronlm-2}. To mitigate this problem, \axonn implements an
intelligent memory optimization algorithm that enables it to reduce the number
of GPUs required to train a single instance of a neural network by four times.
This is made possible by using the CPU memory as a scratch space for offloading
data periodically during training and prefetching it intelligently whenever
required. To extend training to a large number of GPUs, we employ data
parallelism wherein multiple instances of the neural network are trained in an
embarrassingly parallel manner~\cite{pytorchdist-vldb}. When evaluated against
a 12 billion parameter transformer~\cite{transformer} neural network, our
memory optimizations yield a performance improvement of over 13\%.  

We demonstrate the scalability of our implementation and compare its
performance with that of existing state-of-the-art parallel deep learning
frameworks - Megatron-LM~\cite{megatronlm-2} and DeepSpeed~\cite{zero_3D,
sc2020zero}.  Our experiments show that \axonn comprehensively outperforms
other frameworks in both weak scaling and strong scaling settings. When tested
against large GPT-style~\cite{gpt-3} transformer models with 12, 24, 50 and 100
billion parameters on 48, 96, 192 and 384 NVIDIA Tesla V100 GPUs respectively,
\axonn achieves an impressive per-GPU throughput of 49.4--54.78\% of
theoretical peak and reduces the training time by 22-37 days as compared to the
next best framework -- DeepSpeed.  Our framework can thus help researchers save
time and resources in their experiments.

Our contributions can be summarized as follows:
\begin{itemize}
    \item We propose a MPI-based point-to-point communication backend for
parallel deep learning that exploits asynchrony to overlap communication with
computation and thus increase hardware efficiency.
    \item We also implement a message-driven scheduler that enables the
scheduling of neural network operations in the order in which their data
dependencies are met.
    \item We develop a proof that explains the reasons for 
inefficiency of certain parallel deep learning algorithms at scale. We then
propose a novel memory optimization algorithm that mitigates this issue by
using the CPU memory as a scratch space to offload and prefetch data
intelligently.
    \item We present a user-friendly, open-source implementation of
AxoNN\footnote{\url{https://github.com/hpcgroup/axonn}}, which places minimal
programming burden on the end-user, similar to the familiar single GPU
PyTorch programming environment.
\end{itemize}

\section{Background on deep learning}
This section provides a background on training  neural networks, and different
modes of parallelism in deep learning.

\subsection{Definitions and basics of training neural networks}
\label{sec:defs}

We now describe the basic terminology and the layout of a typical neural
network training procedure.

\vspace{0.08in}
\noindent{\bf Neural networks:}
are parameterized function approximators. Their popularity stems from their
flexibility to approximate a plethora of functions on high dimensional input
data. The training algorithm iteratively adjusts the values of the parameters
to fit the input data accurately. We collectively refer to the entire parameter
vector of the neural network as $\vec{\theta}$.

\vspace{0.08in}
\noindent{\bf Layers:}
Computation of a neural network is divided into layers. Each layer consumes the
output of its previous layer. The first layer operates directly on the input.
The output of the last layer is the final output of the neural network. The
outputs of intermediate layer are called activations. We refer to a neural
network with N layers as $(l_{0}, l_{1}, l_{2}, ... l_{N-1})$, where $l_{i}$
stands for the $i^{th}$ layer. $\vec{\theta_{i}}$ and $\vec{a_{i}}$ refer to
the parameters and activations of the $i^{th}$ layer respectively.

\vspace{0.08in}
\noindent{\bf Loss:}
is a scalar which defines how well a given set of parameters of a neural
network $\vec{\theta}$ approximate the input dataset. The task of training is
essentially a search for the value of $\vec{\theta}$, which minimizes the loss
function - $L(X,\vec{\theta})$, where $X$ is the input dataset. 

\vspace{0.08in}
\noindent{\bf Forward and backward pass:}
A neural network first computes the activations for each layer and subsequently
the loss function for a given $(X,\vec{\theta})$. This is called the forward
pass. After that it computes the gradient of the loss w.r.t.~the parameters
i.e. $\nabla\vec{\theta} = \frac{dL(X,\vec{\theta})}{d\vec{\theta}}$ via the
backpropagation algorithm \cite{Rumelhart:1986backprop}. This is called the
backward pass.

\vspace{0.08in}
\noindent{\bf Optimizer:}
The optimizer uses $\nabla\vec{\theta}$ to update $\vec{\theta}$ to a value
that incrementally reduces the value of $L(X,\vec{\theta})$. A training run
includes repeated forward pass, backward passes followed by the optimizer step
to iteratively update $\vec{\theta}$ to a desirable value which fits the data
better. Deep learning optimizers maintain a state vector $\vec{s_{opt}}$ of
size $\mathcal{O}(|\vec{\theta}|)$ which is usually a running history of past
gradient vectors. The updated $\vec{\theta}$ is a function of $\vec{s_{opt}}$
and $\nabla\vec{\theta}$. 

\vspace{0.08in}
\noindent{\bf Batch:}
Training algorithms do not use the entire dataset $X$ for computing the loss.
Instead mutually exclusive and exhaustive subsets called batches of the dataset
are repeatedly sampled for training. The cardinality of a batch is called the
batch size.

\vspace{0.08in}
\noindent{\bf Mixed precision:}
Mixed precision training involves keeping two copies of the network parameters
in single and half precision \cite{micikevicius2018mixed}. Forward and backward
passes are done in half precision to boost performance. However, the optimizer
step is applied to the single precision copy of the weight.  To prevent
underflow during the calculation of half precision gradients, the loss is
typically scaled up by multiplication with a large number called the scaling
factor. The optimizer first converts the half precision gradients into full
precision and then descales them to obtain their true value.  Mixed precision
computation can take advantage of the fast Tensor Cores present in modern
hardware accelerators such as the NVIDIA V100 and A100 GPUs. We refer to half
precision parameters and gradients as $\vec{\theta_{16}}$ and
$\nabla\vec{\theta_{16}}$ respectively and their full precision counterparts as
$\vec{\theta}$ and $\nabla\vec{\theta}$ respectively.

\subsection{Modes of parallelism in deep learning}
\label{sec:modes-par}

Algorithms for parallel deep learning fall into three categories - data
parallelism, intra-layer parallelism and inter-layer parallelism. Frameworks
that rely on more than one kind of parallelism are said to be implementing
hybrid parallelism. 

\vspace{0.08in}
\noindent{\bf Data parallelism:}
In data parallelism, identical copies of the model parameters are distributed
among GPUs. Input batches are divided into equal sized chunks and consumed by
individual GPUs for computation, which proceeds independently on each GPU.
After that a collective all-reduce communication is initiated to average the
parameter gradients on each rank. The average gradients are used to update the
local copies of the network parameters. Due to its embarrassingly parallel
nature, data parallelism scales very efficiently in practice. However, vanilla
data parallelism is restricted by the need to fit the entire model in a single
GPU. To overcome this restriction, it has to be combined with inter-layer and
intra-layer parallelism for training extremely large neural networks such as
GPT-3~\cite{gpt-3}.

\vspace{0.08in}
\noindent{\bf Intra-layer parallelism:}
divides the computation of each layer of the neural network on multiple GPUs.
Each GPU is responsible for partially computing the output activation of a
layer. These partial outputs are pieced together using a collective
communication primitive like all-gather or all-reduce to be used for the
computation of the next layer. For example,
Megatron-LM~\cite{megatronlm,megatronlm-2} shards the various matrix
multiplications of a transformer~\cite{transformer} layer across GPUs. While
saving memory, it is prohibited by expensive collective communication after
computing the output activations. Typically, intra-layer parallelism cannot
scale efficiently beyond the confines of GPUs inside a node connected via a
high-speed inter-connect like NVLink~\cite{megatronlm-2}. 

\begin{figure}[h]
    \centering
      \includegraphics[width=\columnwidth]{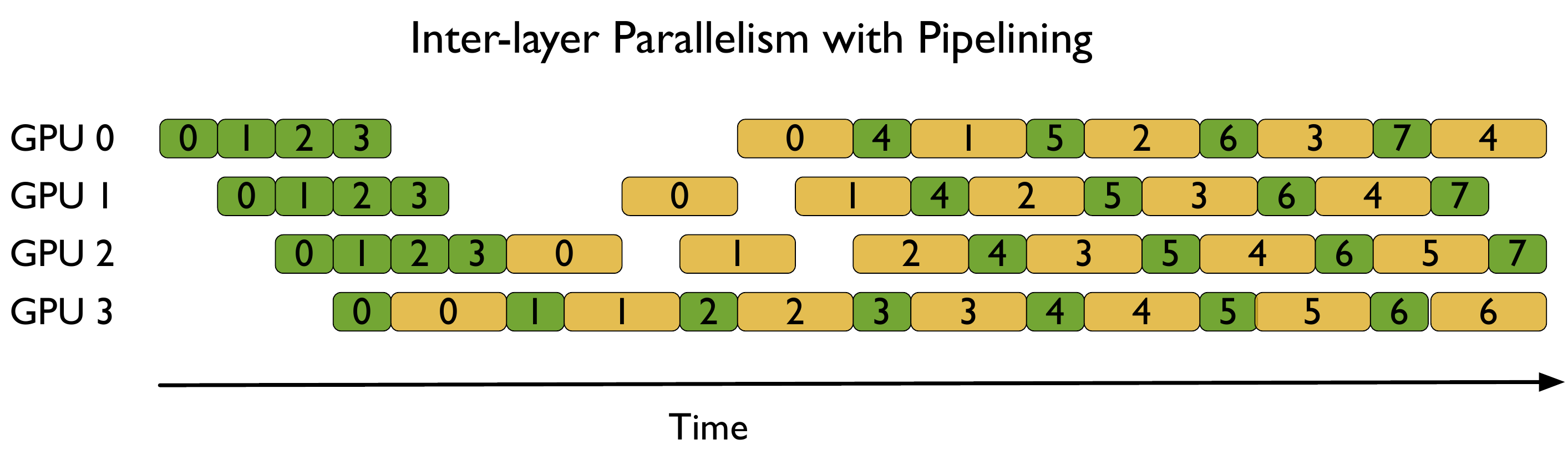}
    \caption{Inter-layer parallelism on four GPUs. The green and yellow
boxes represent the forward and backward passes of a microbatch respectively.
The numbers inside each box represent the microbatch number. We assume that the
backward pass takes twice as much time as the forward pass.}
    \label{fig:inter-layer-parallelism}
\end{figure}

\vspace{0.08in}
\noindent{\bf Inter-layer parallelism:}
divides the layers of a neural network among worker GPUs.  To achieve
parallelism, an input batch is divided into smaller microbatches. Forward and
backward passes for different microbatches can thus proceed on different GPUs
concurrently. Inter-layer parallelism is often called as pipelining and the set
of GPUs implementing it are called the pipeline. Prior work has shown that
inter-layer parallelism is inefficient for a large number of GPUs in the
pipeline due to an increase in the idle time in the
pipeline~\cite{megatronlm-2}. Figure~\ref{fig:inter-layer-parallelism}
illustrates the working of inter-layer parallelism.

\vspace{0.08in}
\noindent{\bf Hybrid parallelism:}
Data parallelism is often combined with either or both of intra-layer or
inter-layer parallelism to realize hybrid parallelism. For example,
Megatron-LM~\cite{megatronlm-2} and DeepSpeed~\cite{sc2020zero,zero_3D} combine
all three forms of parallelism to train large transformer neural networks~\cite
{transformer} at extreme scale. This form of parallelism has been called 3D
parallelism in literature. Prior work~\cite{zero_infinity,megatronlm-2} has
shown 3D parallelism as the fastest method for training large scale neural
networks.

\section{Designing a parallel deep learning framework} 
\label{sec:design}
We now present the design of our new framework. \axonn combines inter-layer
parallelism and data parallelism to scale parallel training to a large number
of GPUs. 

\subsection{A hybrid approach to parallel training}

\begin{figure}[h]
    \centering
      \includegraphics[width=\columnwidth]{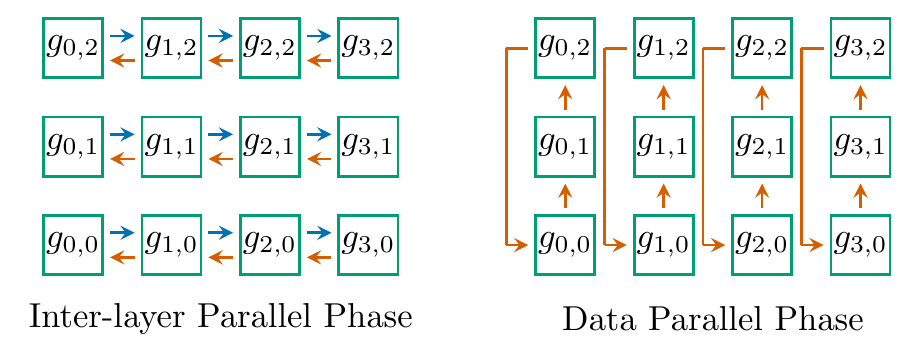}
    \caption{\axonn uses hybrid parallelism that combines inter-layer
and data parallelism. In this example, we train a neural network
on 12 GPUs in a $4\times 3$ configuration (4-way inter-layer parallelism and 3-way data parallelism).
The blue and red arrows represent communication of
activations and gradients respectively. In inter-layer parallelism, these
gradients are w.r.t.~the output activations, whereas in data parallelism, these
gradients are w.r.t.~the network parameters.}
    \label{fig:axonn-design}
\end{figure}

The central idea behind \axonn's hybrid parallelization of neural networks is
to create a hierarchy of compute resources (GPUs) by dividing them into equally
sized groups. Each group of GPUs can be treated as a unit that has a full copy
of the network similar to a single GPU in the case of pure data parallelism.
Each group works on different shards of a batch concurrently to provide data
parallelism. GPUs within each group are used to parallelize the computation
associated with processing a batch shard using inter-layer parallelism.  In the
case of \axonn, we arrange GPUs in a virtual 2D grid topology as shown in
Figure~\ref{fig:axonn-design}. GPUs in each row form a group and are used to
implement inter-layer parallelism within each group. The groups together are
used to provide data parallelism by processing different shards of a batch in
parallel. We use $\mathit{G_{data}}$ and $\mathit{G_{inter}}$ to denote the
number of data-parallel groups and the number of GPUs inside each data-parallel
group respectively.

\begin{algorithm}[h]
    \caption{\axonn's hybrid training algorithm for GPU $g^{i,j}$ in a $G_{\mathit{inter}} \times G_{\mathit{data}}$ configuration } \label{alg:axonn}
    {\small 
    \begin{algorithmic}[1]
    \Function{train}{neural\_network, dataset ...} \label{func:train}
        \State nn\_shard $\gets$ instantiate neural network shard for $g^{i,j}$  
        \While{training has not finished}
            \State next\_batch $\gets$ get next batch from dataset
            \State batch\_shard $\gets$ get batch shard for $g^{i,j}$
            \State \Call{data\_parallel\_step}{nn\_shard, batch\_shard ...}
            \State run the optimizer
        \EndWhile
    \EndFunction

    \State
    \Function{data\_parallel\_step}{nn\_shard, batch\_shard ...}        
        \State \Call{inter\_layer\_parallel\_step}{nn\_shard, batch\_shard ...}
        \State \textcolor{mauve}{All-reduce on nn\_shard.$\nabla\vec{\theta}$}
    \EndFunction
    \end{algorithmic}
    }
\end{algorithm}

Algorithm~\ref{alg:axonn} explains the working of \axonn's parallel algorithm
from the point of view of one of the GPUs $g^{i,j}$ in the 2D virtual grid.
Training begins in the \textproc{train} function (line 1) which takes a neural
network specification and a training dataset as its arguments.  For each GPU,
we first instantiate a neural network shard (contiguous subset of layers) that
GPU $g^{i,j}$ will be responsible for in the inter-layer phase (line 12). In
the main training loop (lines 3-7), we divide the input batch into
$G_{\mathit{data}}$ shards (line 5) and run the data parallel step on the
corresponding shard of $g^{i,j}$. The data parallel step first calls the
inter-layer parallel step followed by an all-reduce on the gradients of the
network shard. In the optimizer phase, we run a standard optimizer used in deep
learning such as Adam~\cite{KingmaAdam2014} as described in
Section~\ref{sec:defs}. Next, we provide details about the inter-layer and
data-parallel phases in \axonn.

\subsection{Data parallel phase}
\label{sec:des:dp}

We outline \axonn's data parallel phase in Algorithm~\ref{alg:axonn}. The
central idea of \axonn's data parallelism is to divide the computation of a
batch by breaking it into $G_{\mathit{data}}$ equal sized shards and assigning
each individual shard to a row of GPUs in Figure~\ref{fig:axonn-design} (line
5).  Each row of GPUs then initiates the inter-layer parallel phase on their
corresponding batch shards (line 12). On completion of the inter-layer parallel
phase, GPUs in a column of Figure~\ref{fig:axonn-design} issue an all-reduce on
the gradients ($\nabla\vec{\theta}$) of their network shards (line 13). This
marks the completion of the data parallel phase, after which we run the
optimizer to update the weights (line 7).

\subsection{Inter-layer parallel phase}
\label{sec:des:ilp}

The algorithm for the inter-layer parallel phase in \axonn is described in
Algorithm~\ref{alg:axonn-ilp}. We first divide the batch shard into equal sized
microbatches (line 2). The size of each microbatch is a user-defined
hyperparameter. We define the $\mathit{pipeline\_limit}$ as the maximum number
of microbatches that can be active in the pipeline. To make sure computation
can concurrently happen on all GPUs we first inject $\mathit{pipeline\_limit}$
number of microbatches into the pipeline (lines 4-6) by scheduling their
forward passes on each of the first GPUs in a row of
Figure~\ref{fig:axonn-design} (line 6). The output of the forward pass is then
communicated to the next GPU (line 7). We call lines 3-9 the warmup phase.
Once, the pipeline has been initialized with enough microbatches, we enter the
steady state of the computation (lines 11-31).

\begin{algorithm}[h]
    {\small 
    \caption{\axonn's inter-layer parallelism for GPU $g^{i,j}$ in a $G_{\mathit{inter}} \times G_{\mathit{data}}$ configuration } \label{alg:axonn-ilp}
    \begin{algorithmic}[1]
    \Function{inter\_layer\_parallel\_step}{nn\_shard, batch\_shard ...}
        \State microbatches $\gets$ divide batch\_shard into microbatches 
        \If{$i = 0$}
            \For{$\_$ in $\mathit{pipeline\_limit}$}
                \State next\_microbatch $\gets$ microbatches.\Call{pop}{ }
                \State output $\gets$ nn\_shard.\Call{Forward}{next\_microbatch}
                \State \textcolor{mauve}{\Call{Send}{output, $g^{i+1,j}$}}
            \EndFor
        \EndIf
        \State
        \While{messages to receive}
            \State \textcolor{mauve}{msg $\gets$ \Call{Receive}{ }}
            \If{msg.source $=$ ${g^{i-1,j}}$}
                \State output $\gets$ nn\_shard.\Call{Forward}{msg}
                \If{$i = n_{inter} - 1$}
                    \State output $\gets$ nn\_shard.\Call{Backward}{1}
                    \State \textcolor{mauve}{\Call{Send}{output, $g^{i-1,j}$}}
                \Else
                    \State \textcolor{mauve}{\Call{Send}{output, $g^{i+1,j}$}}
                \EndIf
            \ElsIf{msg.source $=$ ${g^{i+1,j}}$}
                \State output $\gets$ nn\_shard.\Call{Backward}{msg}
                \If{$i = 0$}
                    \State next\_microbatch $\gets$ microbatches.\Call{pop}{ }
                    \State output $\gets$ nn\_shard.\Call{Forward}{next\_microbatch}
                    \State \textcolor{mauve}{\Call{Send}{output, $g^{i+1,j}$}}
                \Else
                    \State \textcolor{mauve}{\Call{Send}{output, $g^{i-1,j}$}}
                \EndIf
            \EndIf
            
        \EndWhile
    \EndFunction
    \end{algorithmic}
    }
\end{algorithm}

In the steady state, each GPU repeatedly receives messages (line 12) and starts
the computation for a forward or backward pass of the network shard depending
on if the message is received from a GPU before or after it in its row (lines
15 and 21). If the source is $g^{i-1,j}$ (line 13), a forward pass computation
is done using the received message (line 14). We then send the output of the
forward pass to GPU $g^{i+1,j}$ (line 19) unless GPU $g^{i,j}$ is the last GPU
in the pipeline (line 15). If GPU $g^{i,j}$ is the last GPU in the pipeline, it
initiates the backward pass.  Similarly if the source of the message is
$g^{i+1,j}$ (line 21), the GPU starts the backward pass computation. Once that
is complete, we send the output to $g^{i-1,j}$ (line 28) if GPU $g^{i,j}$ is
not the first GPU in the pipeline. If it is the first GPU, we inject a new
microbatch into the pipeline by initiating its forward pass (lines 24-26).
This ensures that the pipeline always has a steady number of microbatches equal
to the $pipeline\_limit$ in its steady state. This process repeats until all of
the messages for all microbatches of the batch shard have been received and
processed (line 11).

\section{Implementation of \axonn}
\label{sec:implementation}
In this section, we provide details of the implementation of \axonn in Python
using MPI, NCCL~\cite{nccl}, and PyTorch~\cite{paszke2019pytorch-nips}.  \axonn
is designed to be run on GPU-based clusters ranging from a single node with
multiple GPUs to a large number of multi-GPU nodes.  Following the MPI model,
\axonn launches one process to manage each GPU. Each process is responsible for
scheduling communication and computation on its assigned GPU. We use PyTorch
for implementing and launching computational kernels on the GPU. \axonn relies
on mixed-precision trainingfor improved hardware
utilization~\cite{micikevicius2018mixed}.

GPUs consume data at a very high rate. As an example, the peak half-precision
performance of V100 GPUs on Summit is a staggering 125 Tflop/s. Ensuring that
the GPUs constantly have data to compute on requires designing an efficient
inter-GPU communication backend, both for inter-layer and data parallelism.  We
use NVIDIA's GPUDirect technology, which removes redundant copying of data to
host memory and thus decreases the latency of inter-GPU communication. We use
CUDA-aware MPI for point-to-point communication in the inter-layer parallel
phase. In the data parallel phase, we use NCCL for collective communication.
We provide explanations for our choices in the sections below.

\subsection{Inter-layer parallel phase}
\label{imp:inter-layer}

We first fix the $\mathit{pipeline\_limit}$ to $G_{\mathit{inter}}$ for our
implementation of inter-layer parallelism. This ensures that all the GPUs in
the pipeline can compute concurrently while placing a low memory overhead for
storing activations. When implementing the point-to-point sends and receives in
Algorithm~\ref{alg:axonn}, we had a choice between CUDA-aware MPI and NVIDIA's
NCCL library, both of which invoke GPUDirect for inter-GPU communication.  We
compared the performance of the two libraries for point-to-point and collective
operations on \summit using the OSU Micro-benchmarks v5.8~\cite{osu-5.8}.

\begin{figure}[h]
    \centering
      \includegraphics[width=0.95\columnwidth]{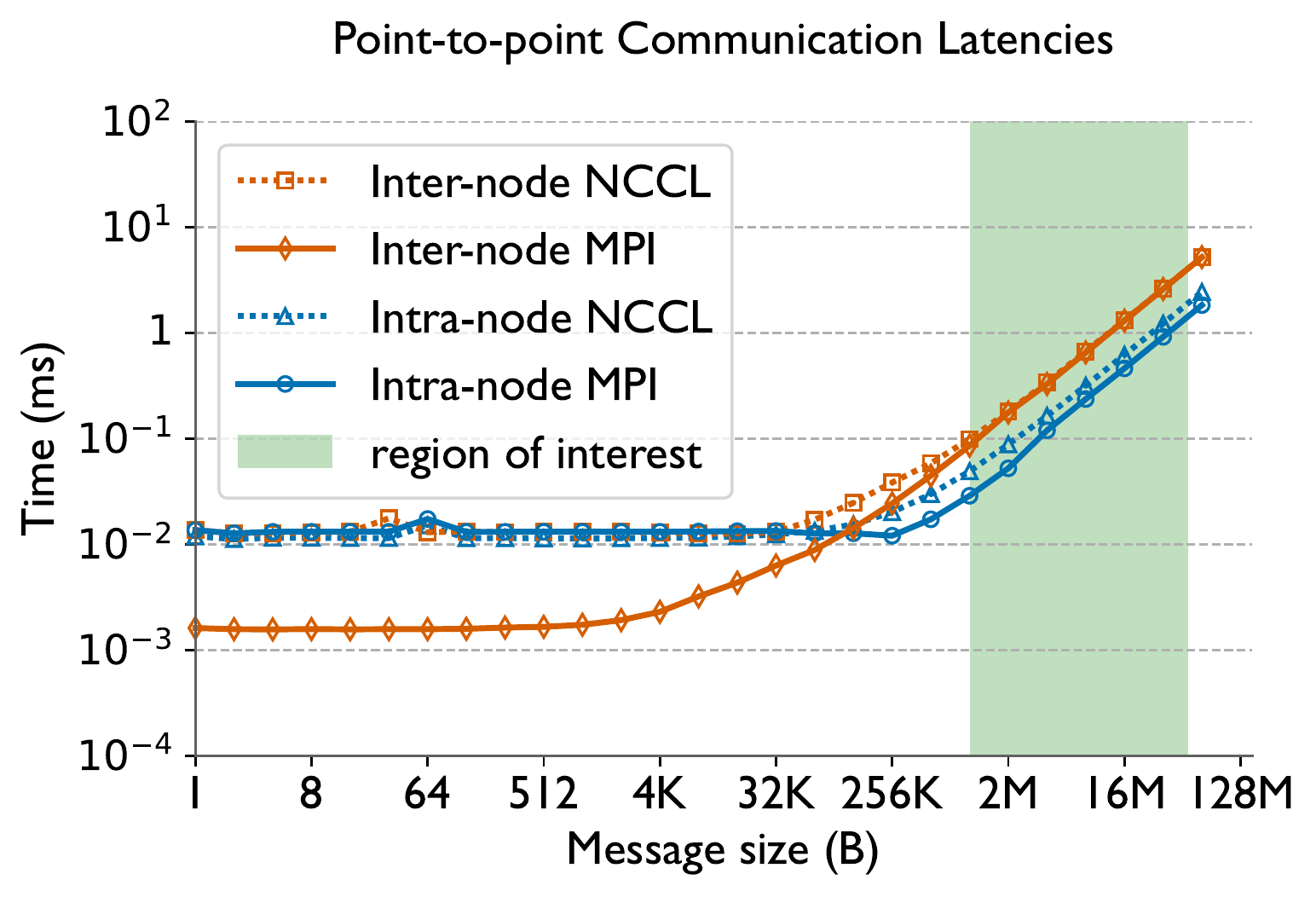}
    \caption{Execution time for point-to-point messages in MPI and NCCL on Summit using the OSU Micro-benchmarks v5.8~\cite{osu-5.8}. We use two GPUs on the same and different nodes for the intra- and inter-node experiments respectively.}
    \label{fig:p2p}
\end{figure}

Figure~\ref{fig:p2p} compares the performance of MPI and NCCL for intra-node
(GPUs on the same node) and inter-node (GPUs on different nodes) point-to-point
messages (the \texttt{osu\_latency} ping pong benchmark).  Typical sizes of
messages exchanged during point-to-point communication in deep learning
workloads are in the range of 1--50 MB. In Figure~\ref{fig:p2p}, we see nearly
identical inter-node latencies. However, MPI clearly outperforms NCCL for
intra-node communication. Further, NCCL point-to-point communication primitives
block on the communicating GPUs until a handshake is completed.  MPI on the
other hand allows the user to issue sends/receives without blocking other
computation kernels on the GPU. We thus implement \axonn's asynchronous
point-to-point communication backend using MPI.

We build our implementation of inter-layer parallelism on top of
MPI4Py~\cite{mpi4py}, a library which provides Python bindings for the MPI
standard. We use \texttt{MPI\_Isend} and \texttt{MPI\_Irecv} to implement the
\textproc{Send} and \textproc{Receive} methods mentioned in
Algorithm~\ref{alg:axonn}. The \texttt{MPI\_Irecv}~s are issued preemptively at
the beginning of a forward or backward pass to overlap the reception of the
next message with the computation and thus achieve asynchronous messaging. In
lines 13 and 21 of Algorithm~\ref{alg:axonn-ilp}, we have already shown how
\axonn is designed to support message driven scheduling for inter-layer
parallelism. Combined with the asynchronous point-to-point communication
backend discussed in this section, we are able to realize our goal of
asynchronous message driven scheduling for improving hardware utilization. 

\subsection{Data parallel phase} \label{imp:data}

\begin{figure}[h]
    \centering
      \includegraphics[width=0.95\columnwidth]{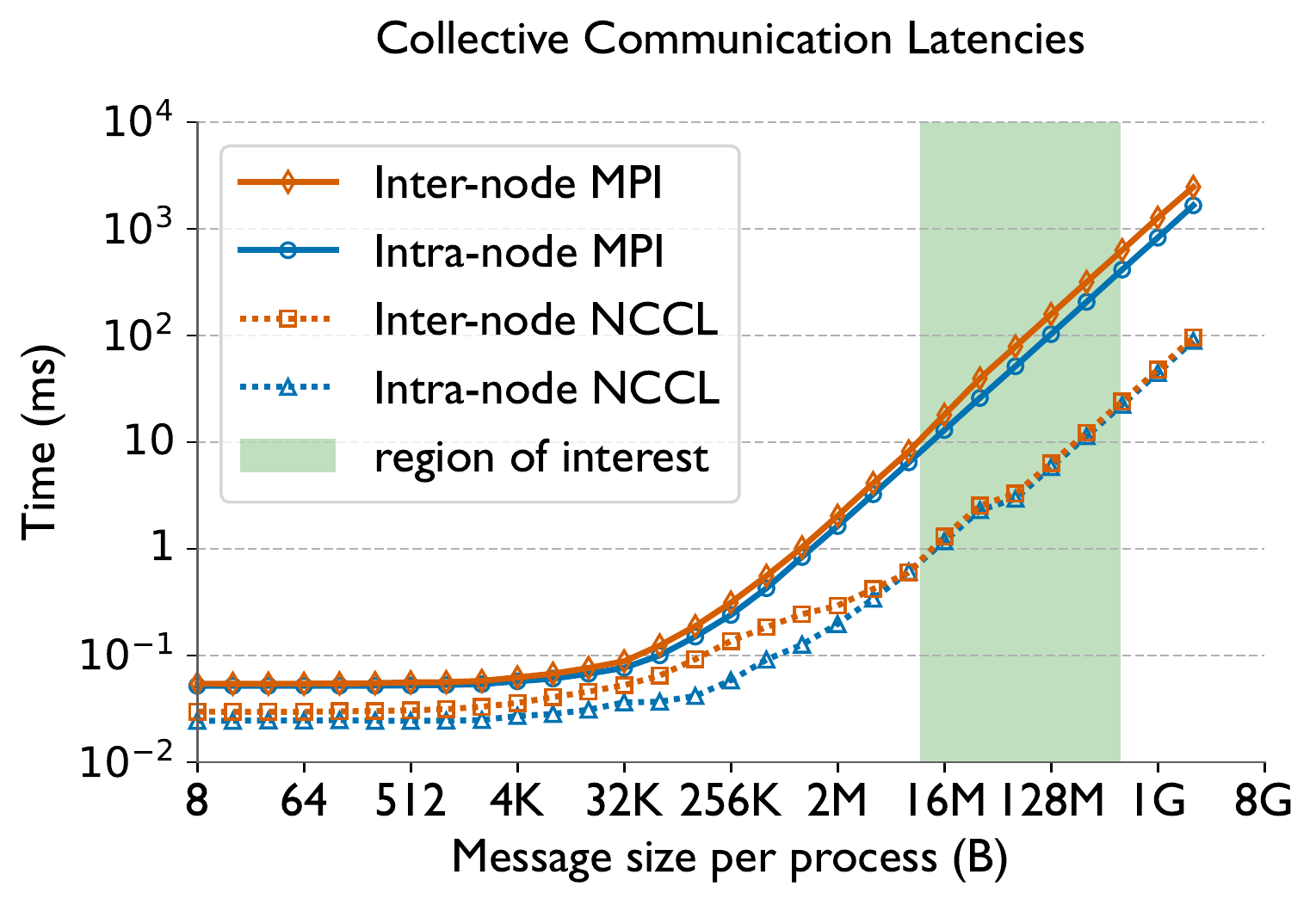}
    \caption{Execution time for an All-reduce operation in MPI and NCCL on
Summit using the OSU Micro-benchmarks v5.8~\cite{osu-5.8}. We use six and
twelve GPUs for the intra-node and inter-node experiments respectively.}
    \label{fig:collective}
\end{figure}

We again had a choice between MPI and NCCL for the all-reduce operation in the
data-parallel phase and we decided to make that choice based on empirical
evidence.  Figure~\ref{fig:collective} presents the performance of the
all-reduce operation using MPI and NCCL. In this case, intra-node refers to
performing the collective over all six GPUs of a single node and inter-node
refers to performing it over 12 GPUs on two nodes. The results demonstrates the
significantly better performance of NCCL (dashed lines) over MPI for collective
communication. This makes NCCL the clear choice for \axonn's collective
communication backend.

Our implementation of data parallelism uses NCCL for the all-reduce operation
over half-precision parameter gradients. We avoid using full-precision
gradients to reduce communication times. To prevent overflow in the all-reduce
operation, we pre-divide the loss by the total number of microbatches in the
input batch. We use PyTorch's \texttt{torch.distributed}
API~\cite{pytorchdist-vldb} for making NCCL all-reduce function calls.

\section{Memory and performance optimizations}
In this section, we discuss optimizations that are critical to the memory
utilization and performance of \axonn. We implement these optimizations on top
of the basic version of our framework discussed in
Section~\ref{sec:implementation}.  We verify their efficacy by conducting
several experiments on a 12 billion parameter transformer neural
network~\cite{transformer} on 48 NVIDIA V100 GPUs.  Section~\ref{sec:setup:nn}
provides the exact architectural hyperparameters of this model in
Table~\ref{tab:weak-scaling}, and describes how these hyperparameters affect
the computational and memory requirements of a transformer.  For all the
experiments in this section, the batch size and sequence length are fixed at
2048 and 512 respectively, and we use mixed-precision
training~\cite{micikevicius2018mixed} with the Adam
optimizer~\cite{KingmaAdam2014}.  

\subsection{Memory optimizations for reducing activation memory}
\label{sec:act-ckp}

Gradient checkpointing reduces the amount of memory used to store activations
by only storing the output activations of a subset of layers during the forward
pass~\cite{chen2016training} . For a neural network with N layers $(l_{1},
l_{2}, .. l_{N})$, this subset of layers is defined as $S_{\mathit{ckp}} =
(l_{\mathit{ac}}, l_{2\cdot \mathit{ac}}, l_{3\cdot \mathit{ac}}.... l_{N})$,
where $\mathit{ac}$ is a tunable hyperparameter with the constraint that it
should be a factor of N.  Activations for layers that are not checkpointed are
regenerated during their backward pass. For inter-layer parallelism, it can be
shown that the maximum activation memory occupied per GPU with $\mathit{ac}$ as
the gradient checkpointing hyperparameter is:
\begin{equation}
\mathrm{M_{activation}} \propto G_{\mathit{inter}} \times \left ( \frac{N}{G_{\mathit{inter}} \times \mathit{ac}} \right ) + 1 + ac 
\end{equation}

The value $\mathit{ac} = \sqrt{N}$ leads to the lowest value of
$\mathrm{M_{activation}}$. We thus set the value of $ac$ to the factor of
$\frac{N}{G_{\mathrm{inter}}}$ (the number of layers on each GPU) closest to
$\sqrt{N}$. To the best of our knowledge, our work is the first to derive an
optimal value of this hyperparameter for inter-layer parallelism. We implement
gradient checkpointing using the \texttt{torch.utils.checkpoint} API provided
by PyTorch~\cite{paszke2019pytorch-nips}. 

\subsection{Memory optimizations for reducing $G_{\mathit{inter}}$ and improving performance of the inter-layer parallel phase}
\label{sec:pipe-cont}

Narayanan et al.~show empirically that the performance of inter-layer
parallelism deteriorates with increasing values of
${G_\mathit{inter}}$~\cite{megatronlm-2}. They attribute this to an increase in
the idle time in the warmup phase with increasing ${G_\mathit{inter}}$. Below,
we show analytically that it is not only the warmup phase, but the entire
inter-layer parallel phase which suffers from inefficiency with increasing
${G_\mathit{inter}}$ by virtue of rising communication to computation ratios. 

\vspace{0.08in}
\begin{lemma}
    \label{lemma:comm}
\emph{Given a neural network, batch size, and number of GPUs, the total amount of communication per GPU in the inter-layer parallel phase is proportional to $\mathit{G_{inter}}$.}
\end{lemma}

\vspace{0.05in}
\noindent{\bf Proof:} The total amount of input data a GPU computes on reduces
as the number of GPUs used for data parallelism increases. Hence, the amount of
data per GPU is inversely proportional to $\mathit{G_{data}}$ given a fixed
total number of GPUs that are split between data and inter-layer parallelism.
It also follows that the amount of data computed on per GPU is directly
proportional to $\mathit{G_{inter}}$. Assuming that the output activations of
each layer of the neural network have the same size, the amount of
point-to-point communication per GPU only depends on the amount of input data
it consumes and not on the number of layers it houses. Thus the total amount of
communication is directly proportional to $\mathit{G_{inter}}$.

\vspace{0.08in}
\begin{lemma}
    \label{lemma:comp}
\emph{Given a neural network, batch size, and number of GPUs, the total amount of computation per GPU in the inter-layer parallel phase is independent of $\mathit{G_{inter}}$.}
\end{lemma}

\vspace{0.05in}
\noindent{\bf Proof:} We have shown in Lemma~\ref{lemma:comm} that the total
amount of input data a GPU computes on is directly proportional to
$\mathit{G_{inter}}$. Since layers are sharded evenly among the
$\mathit{G_{inter}}$ GPUs of a row in Figure~\ref{fig:axonn-design}, the total
amount of computation per instance of the input is inversely proportional to
$\mathit{G_{inter}}$. Thus the total amount of computation per GPU is
independent of $\mathit{G_{inter}}$.

\vspace{0.08in}
\begin{theorem}
    \label{theorem:ginter}
\emph{Given a neural network, batch size, and number of GPUs, the ratio of the total amount of computation to communication in the inter-layer parallel phase is inversely proportional to $\mathit{G_{inter}}$.}
\end{theorem}

\vspace{0.05in}
\noindent{\bf Proof:} Directly follows from Lemmas~\ref{lemma:comm}
and~\ref{lemma:comp}.

\begin{figure}[h]
    \centering
      \includegraphics[width=0.95\columnwidth]{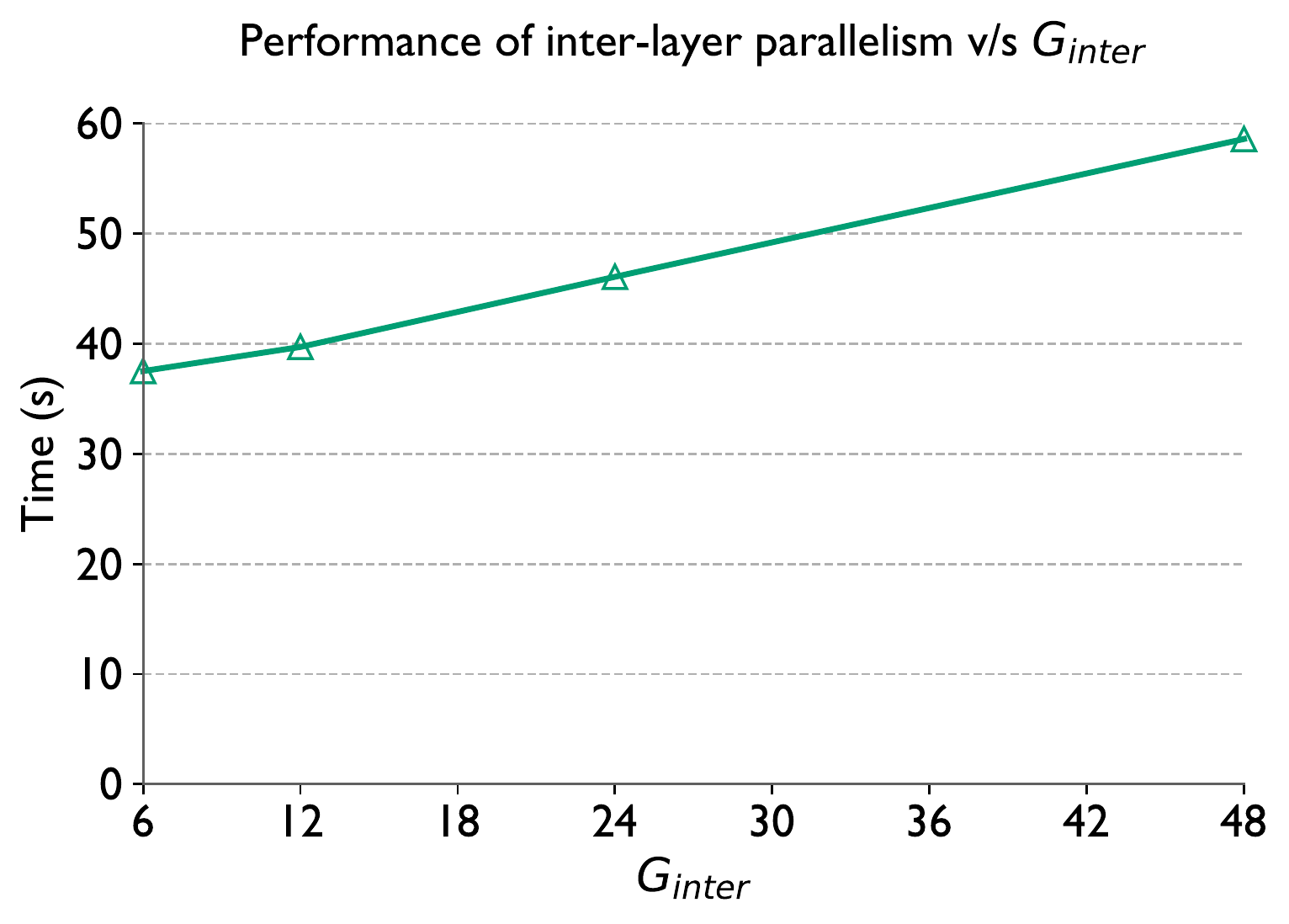}
    \caption{Time spent in the inter-layer parallel phase for a single batch for different values of ${G_\mathit{inter}}$ when training a 12 billion parameter transformer model on 48 GPUs of Summit.}
    \label{fig:pipeline-depth}
\end{figure}

\vspace{0.08in}
We thus expect the efficiency of inter-layer parallelism to decrease with
increasing $\mathit{G_{inter}}$ due to an increase in the ratio of
communication to computation. We empirically verify this phenomenon by studying
the effect of varying $G_\mathit{inter}$ on performance. We gather the time
spent in the inter-layer parallel phase for processing a batch of input using
our reference transformer neural network for values of $G_\mathit{inter} = [6,
12, 24, 48]$. The corresponding value of $G_\mathit{data}$ is automatically set
to be $\frac{48}{G_\mathit{inter}}$.  We fix the microbatch size and batch size
to 1 and 2048 respectively. We also remove the optimizer states so that we do
not run out of memory for lower values of $G_\mathit{inter}$.
Figure~\ref{fig:pipeline-depth} illustrates the results of our experiment. As
expected, we observe significant gains in performance with decreasing
$G_\mathit{inter}$.

Theorem \ref{theorem:ginter} thus provides a motivation to optimize the
implementation by reducing the number of GPUs used for inter-layer parallelism.
However, a smaller value of $G_\mathit{inter}$ requires the entire network to
fit on a smaller number of GPUs, which increases the number of parameters per
GPU. This increases the memory requirements per GPU. Lets say the number of
parameters per GPU is $ \phi = |\vec{\theta}|$ and we are using the Adam
optimizer~\cite{KingmaAdam2014} which stores two state variables per parameter.
The memory required to store model parameters, gradients and the optimizer
states comes out to be $20\phi$ ($4\phi$ bytes each for $\vec{\theta}$ and
$\nabla\vec{\theta}$, $2\phi$ bytes each for $\vec{\theta_{16}}$ and
$\nabla\vec{\theta_{16}}$, and $8\phi$ bytes for $\vec{s_{opt}}$). Note that
this analysis does not include memory required fori storing activations.

At $G_\mathit{inter} = 6$ on our reference transformer, this would amount to 40
GB memory per GPU which is 2.5 times more than the 16 GB DRAM capacity of
Summit's V100 GPUs. To solve this problem, we introduce a novel memory
optimization algorithm that reduces the amount of memory required to store the
model parameters and optimizer states by five times. 

\vspace{0.08in}
\noindent{\bf Implementation:}
In our memory optimization, only the half precision model parameters
($\vec{\theta_{16}}$) and gradients ($\nabla\vec{\theta_{16}}$) reside on the
GPU.  Everything else is either moved to the CPU ($\vec{s_{opt}}$ and
$\vec{\theta}$) or deleted entirely ($\nabla\vec{\theta}$) before the training
begins. The training procedure requires $\vec{s_{opt}}$ and $\vec{\theta}$ on
the GPU in the optimizer phase. We save memory by not fetching the entire
$\vec{\theta}$ and $\vec{s_{opt}}$ arrays to the GPU, but only small equal
sized chunks at a time. We call these chunks buckets and their size as the
bucket-size ($\mathit{bsize}$). After fetching a bucket of $\vec{s_{opt}}$ and
$\vec{\theta}$, we run the optimizer step on this data and offload it back to
the CPU. GPU Memory is saved by reusing buffers across buckets. 

The total memory footprint of the optimizer is only $16\mathit{bsize}$ now
($4\mathit{bsize}$ and $8\mathit{bsize}$ for the $\vec{\theta}$ and
$\vec{s_{opt}}$ buckets respectively and another $4\mathit{bsize}$ for
descaling the half precision gradients). With our memory optimizations, the
total memory requirements to store model parameters and optimizier states is
now $4\phi + 16\mathit{bsize}$, down from $20\phi$. As $bsize \ll \phi$, this
amounts to a 5$\times$ saving in memory utilization. Since the activation
memory is unaffected the total memory saved should obviously be less than this
number. With $G_\mathit{inter}=24$, $G_\mathit{data}=2$, microbatch size 1 and
$bsize=16$ million our total memory usage for the reference transformer reduces
four fold in practice - from 520 GB to 130.24 GB.

\begin{figure}[h]
    \centering
      \includegraphics[width=0.95\columnwidth]{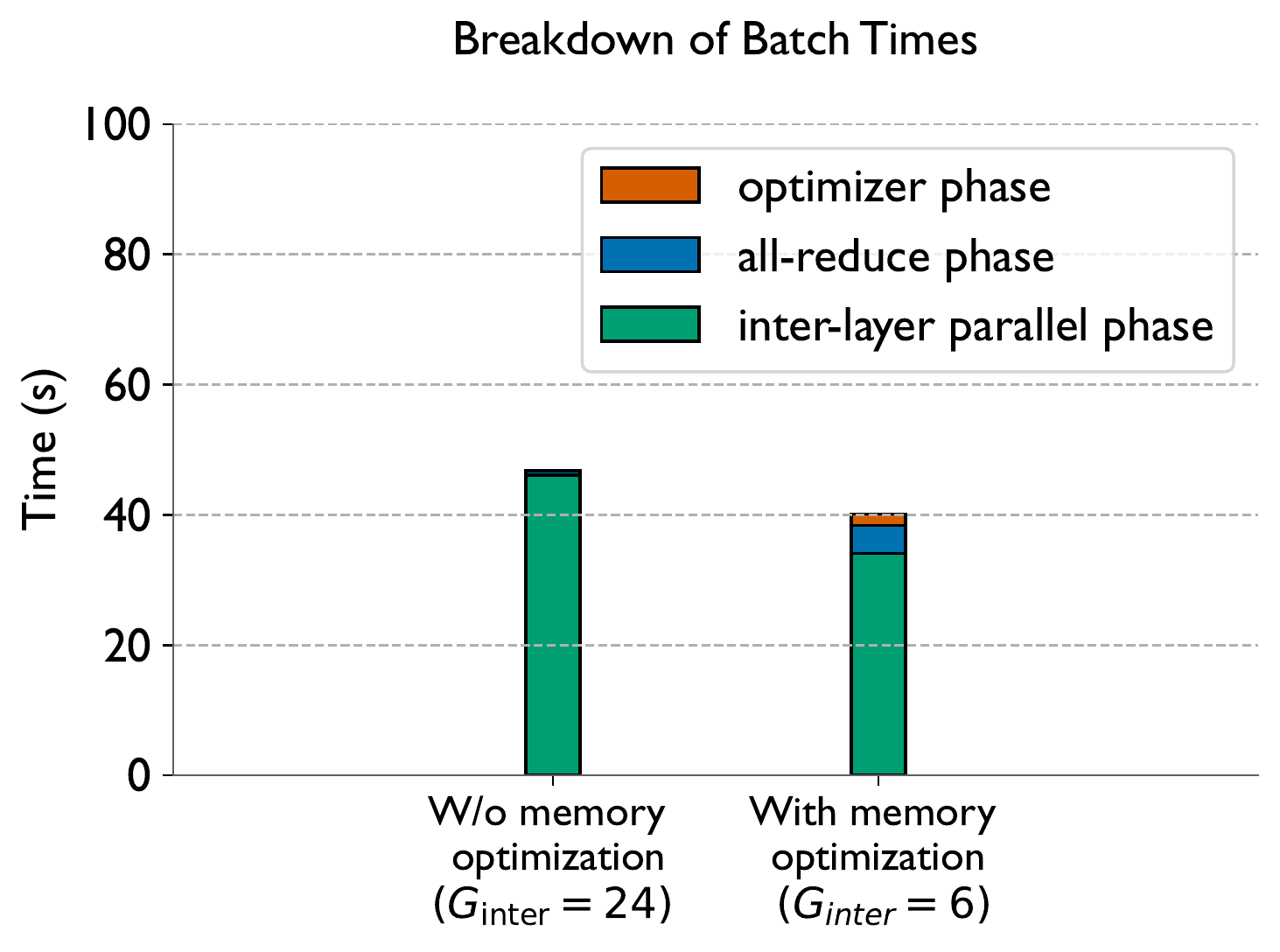}
    \caption{\axonn's performance for a single batch with and without our memory optimization on a 12 billion parameter transformer on 48 GPUs}
    \label{fig:pipeline-contract}
\end{figure}

Next, we use our memory optimization algorithm with $bsize=16$ million to
reduce $G_\mathit{inter}$ from 24 to 6, and study the performance implications.
We expect an increase in the time it takes to complete the data parallel phase,
because both the amount of data and number of GPUs participating in the
all-reduce increases four fold.  Figure~\ref{fig:pipeline-contract} compares
\axonn's performance with and without the memory optimizations. We notice an
improvement of 13 percent in the absolute batch timings. While the time for the
data parallel phase increases from 0.62s to 4.32s, the corresponding
performance gain in the pipelining phase (46.08s v/s 34.05s) compensates for
this increase. We expect higher speedups for larger values of batch size, which
are typical in large scale model training.

\subsection{Overlapping all-reduce \& optimizer phases for performance}

After optimizing the inter-layer parallel phase, we turned our attention to the
less time consuming all-reduce and optimizer phases. We observed that the
all-reduce phase (in blue) takes 2.5 times longer than the optimizer phase (in
green) (right bar in the Figure~\ref{fig:pipeline-contract} plot). We
hypothesize that by interleaving their executions, we could overlap data
movement between the CPU and the GPU in the optimizer phase with the expensive
collective communication of the data parallel phase.  We explain our approach
for enabling this overlap below.

\begin{figure}[h]
    \centering
      \includegraphics[width=\columnwidth]{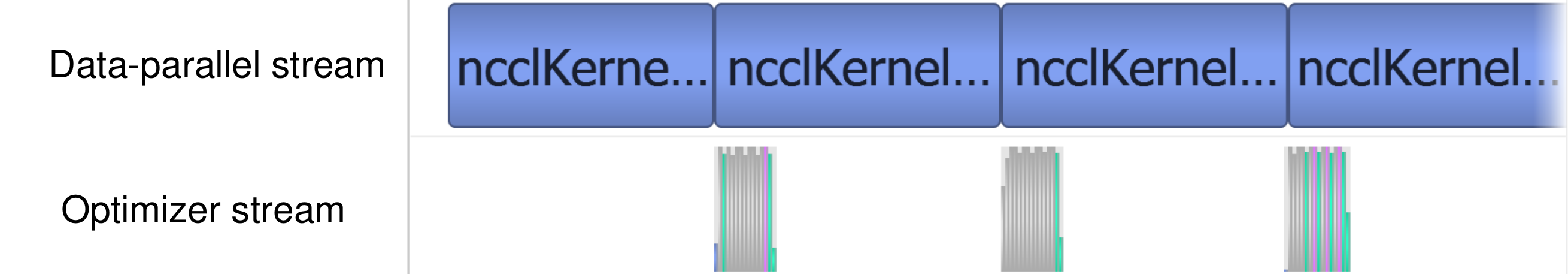}
    \caption{An Nsight profile of \axonn training a 12 billion parameter
transformer model on 48 GPUs shows the interleaving of the all-reduce and optimizer
phases for a single batch. The two rows represent separate CUDA streams for the optimizer and
all-reduce.}
    \label{fig:nsys-optim-dp-overlap}
\end{figure}

\vspace{0.08in}
\noindent{\bf Implementation:}
The main idea here is to issue the all-reduce call into smaller operations over
chunks of the half precision gradients ($\nabla\vec{\theta_{16}}$). For
convenience, we keep the size of the chunk as $k\times\mathit{bsize}$, where we
call $k$ as the all-reduce coarsening factor. As soon as an all-reduce on a
chunk finishes, we enqueue the optimizer step for the corresponding $k$ buckets
and start the all-reduce of the next chunk. The key to achieving overlap is to
use separate CUDA streams for the optimizer and the all-reduce.
Figure~\ref{fig:nsys-optim-dp-overlap} shows an Nvidia Nsight Sytems profile of
our implementation. 

\begin{figure}[h]
    \centering
      \includegraphics[width=0.95\columnwidth]{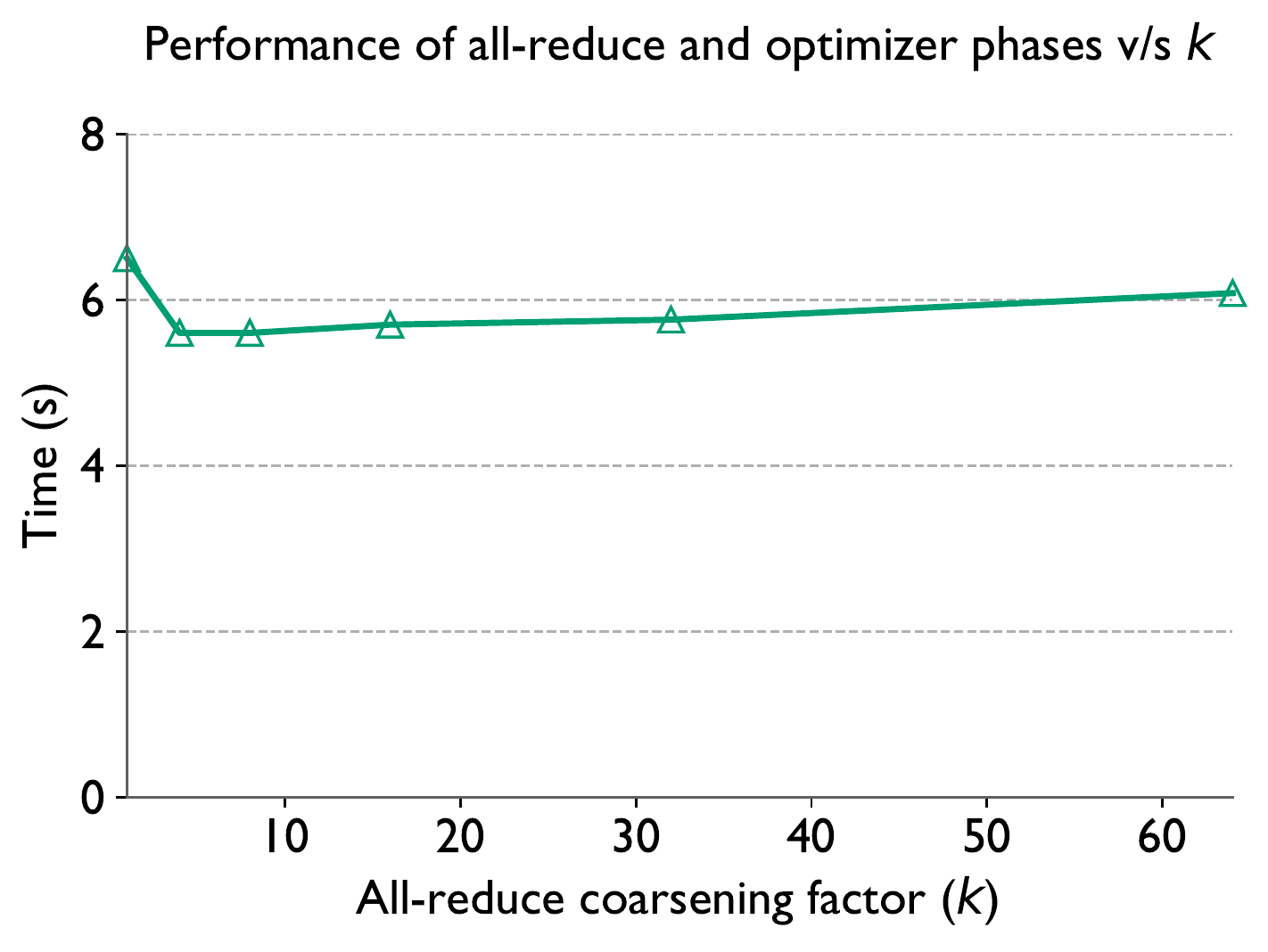}
    \caption{Combined execution time of optimizer and all-reduce phases for a single batch versus the coarsening factor, $k$, for the all-reduce.}
    \label{fig:arc-factor}
\end{figure}

We study the variation of the time it takes to finish the combined data
parallel and optimizer phases with $k$ in Figure~\ref{fig:arc-factor}.  At
$k=1$, we observe high overheads due to too many all-reduce calls.  Infact,
performance is even worse than the case where we had no overlap between the two
phases.  We observe optimum behavior at two and four. Beyond that, we encounter
increasing latencies since increasing $k$ makes the algorithm gravitate towards
sequential behavior.

\section{Experimental setup}
This section provides an overview of our empirical evaluation of \axonn against
the current state-of-the-art frameworks in parallel deep learning. Along with
comparing performance, we also verify the correctness of our implementation by
training a neural network to completion and reporting the loss curves. We
conduct all of our experiments on Oak Ridge National Laboratory's Summit
Cluster. Each node of Summit consists of two Power 9 CPUs each connected to 3
NVIDIA V100 GPUs via NVLink interconnects. The peak intra and inter-node
bandwidth for GPU communication is 50 GB/s and 12.5 GB/s respectively.  Each
V100 GPU has 16 GB DRAM and a peak half precision throughput of 125 Tflop/s.

\subsection{Choice of frameworks}
\label{sec:setup:frameworks}

We compare \axonn with two frameworks implementing 3D parallelism (intra-layer,
inter-layer, and data parallelism), namely - Megatron-LM~\cite{megatronlm-2}
and DeepSpeed~\cite{zero_3D, sc2020zero}, both of which have successfully
demonstrated impressive performance when scaled to models with as many as
trillion parameters. Both these frameworks augment Shoeybi et al.'s intra-layer
parallelism~\cite{megatronlm} with a NCCL based implementation of inter-layer
parallelism. Additionally, DeepSpeed uses the ZeRO~\cite{sc2020zero} family of
memory optimizations which distribute optimizer states across data parallel
GPUs.

\subsection{Choice of neural networks}
\label{sec:setup:nn}

We conduct our strong and weak scaling experiments on GPT-like~\cite{gpt-2,
gpt-3} transformer~\cite{transformer} neural networks on the task of causal
language modeling. For a fair comparison, we use Megatron-LM's extremely
efficient implementation of the transformer kernel for all the three
frameworks. A transformer can be parameterized by three hyperparameters -
number of layers, hidden size, and number of attention heads.  For more details
about the transformer architecture, we refer the reader to Vaswani et
al.~\cite{transformer}.  We first verify the correctness of our implementation
by training GPT-2 small~\cite{gpt-2} (110 million parameters)- to completion.
Table~\ref{tab:weak-scaling} lists the transformer models and the corresponding
GPU counts used in our weak scaling runs.  We start with a 12 billion parameter
transformer on 48 GPUs (8 nodes) and scale up to a 100 billion parameter
transformer on 384 GPUs (64 nodes). We choose 48 GPUs as the starting point as
it was the least number of GPUs the three frameworks could all train the 12
billion parameter transformer without running out of memory. For strong
scaling, we choose the 12 billion parameter transformer from
Table~\ref{tab:weak-scaling} and vary the number of GPUs from 48 to 384. 

\begin{table}[h]
\centering
\caption{\label{tab:weak-scaling}Details about the transformers models used in the weak scaling study}
    \begin{tabular}{rrrccc} \toprule
          &      & Parameters    &        & Hidden    & Attention \\
    Nodes & GPUs & (in Billions) & Layers & Dimension & Heads \\ \midrule
    8     & 48   & 12                                                                  & 48     & 4512                                                        & 24                                                         \\
    16    & 96   & 24                                                                  & 48     & 6336                                                        & 36                                                         \\
    32    & 192  & 50                                                                  & 96     & 6528                                                        & 48                                                         \\
    64    & 384  & 100                                                                 & 96     & 9360                                                        & 60                                                         \\ \bottomrule
    \end{tabular}
\end{table}

\subsection{Dataset and hyperparameters}

The batch size and number of parameters are the two most important quantities
that affect hardware performance. We thus perform two separate experiments that
vary the batch size and number of parameters independently with increasing GPU
counts. Since the dataset size is fixed, both these experiments neatly
translate to a strong scaling and weak scaling setup. We note that it is
absolutely imperative not to vary both these quantities together, otherwise
they can artificially inflate performance numbers. Thus we fix the batch size
for the weak scaling run at 16384. For the strong scaling experiments, we vary
the batch size linearly, starting with 4096 for 48 GPUs (8 nodes) and scaling
upto 32768 for 384 GPUs (48 nodes). We train all our models on the
wikitext-103~\cite{wikitext-103} dataset which consists of around 100 million
English words sourced from more than 28000 Wikipedia articles. We fix the
sequence length and vocabulary size at 512 and 51200 respectively. We use the
Adam optimizer~\cite{KingmaAdam2014} with learning rate 0.001, $\beta_{1}=0.9$,
$\beta_{2}=0.999$ and 0.01 as the decoupled weight decay
regularization~\cite{adamw} coefficient. We tune $G_{\mathit{inter}}$,
$G_{\mathit{intra}}$, and $G_{\mathit{data}}$ for Megatron-LM and DeepSpeed.
For \axonn we just tune $G_{\mathit{inter}}$ and $G_{\mathit{data}}$ since it
does not implement intra-layer parallelism. For AxoNN's memory optimizations we
fix $bsize$ to 4 million and $k$ to 4.

\subsection{Metrics}

We use two metrics in our experiments - namely expected training time and the
percentage of peak half precision throughput.  Both of these are metrics
derived from the average batch time, which we calculate by training for eleven
batches and averaging the timings of the last ten. In accordance with the
training regime employed for GPT-3~\cite{gpt-3}, we define the expected
training time as the total time it would take to train a transformer on a total
of 300 billion tokens.  Let $b$, $s$, $l$, $h$, $V$, $t$ be the batch size,
sequence length, number of layers, hidden size, vocabulary size, and the
average batch time respectively. Then the estimated training time can be
estimated from the batch time as follows:
\begin{equation}
\mathit{estimated\_training\_time} = 3e10^{11}\frac{t}{bs}
\end{equation}
We derive the average flop/s by using Narayanan et al.'s lower bound for flops
in a batch for a transformer~\cite{megatronlm-2}:
\begin{equation}
\mathit{flop/s} = \frac{96bslh^2}{t} (1 + \frac{s}{6h} + \frac{V}{16lh})
\end{equation}
Dividing this by the total peak half precision throughput of all the GPUs used
for training (on Summit this is 125 Tflop/s per GPU) yields the percentage of
peak throughput.

\section{Results}
\begin{figure*}[t]
    \centering
      \includegraphics[width=\columnwidth]{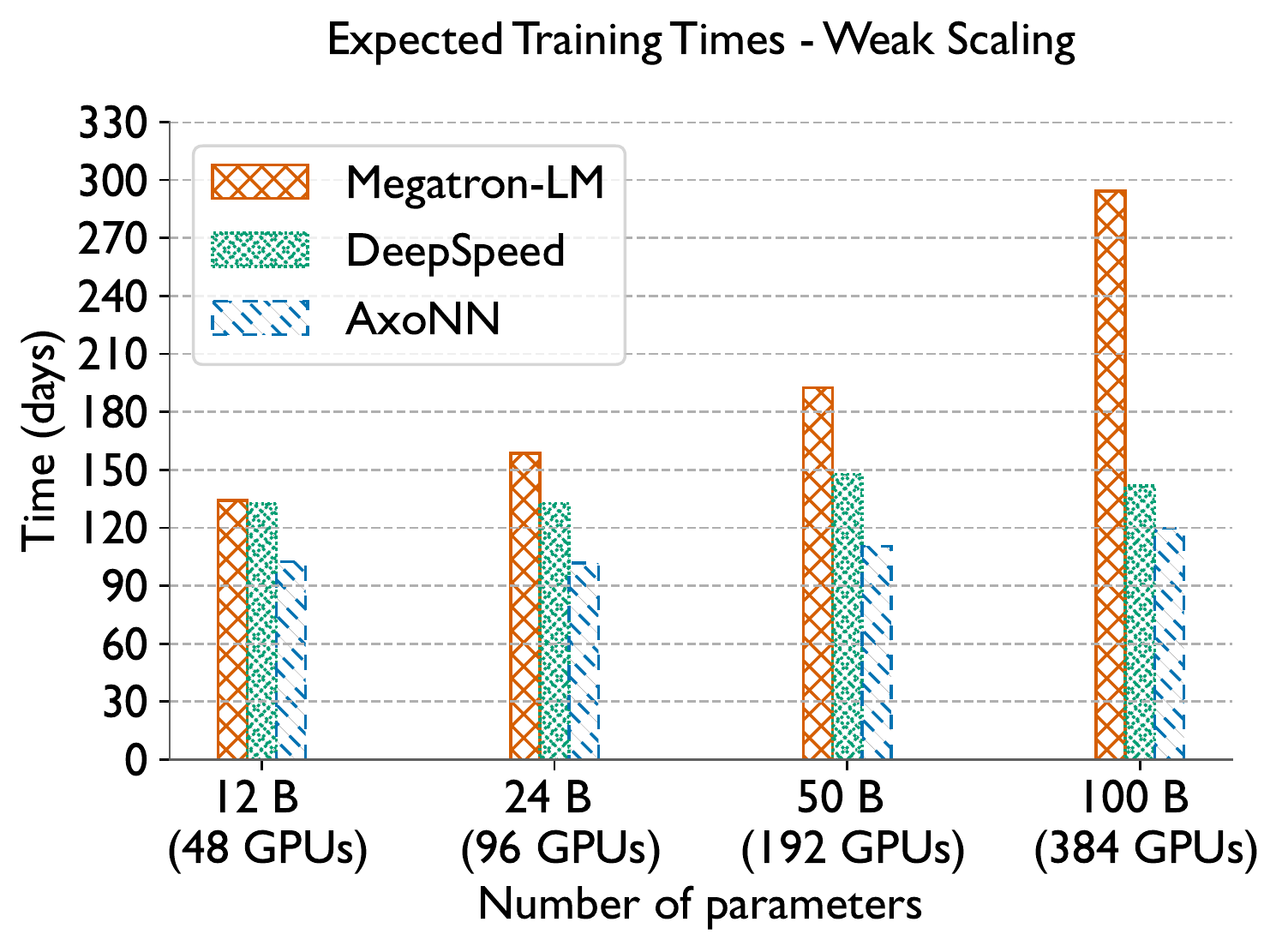}
      \includegraphics[width=\columnwidth]{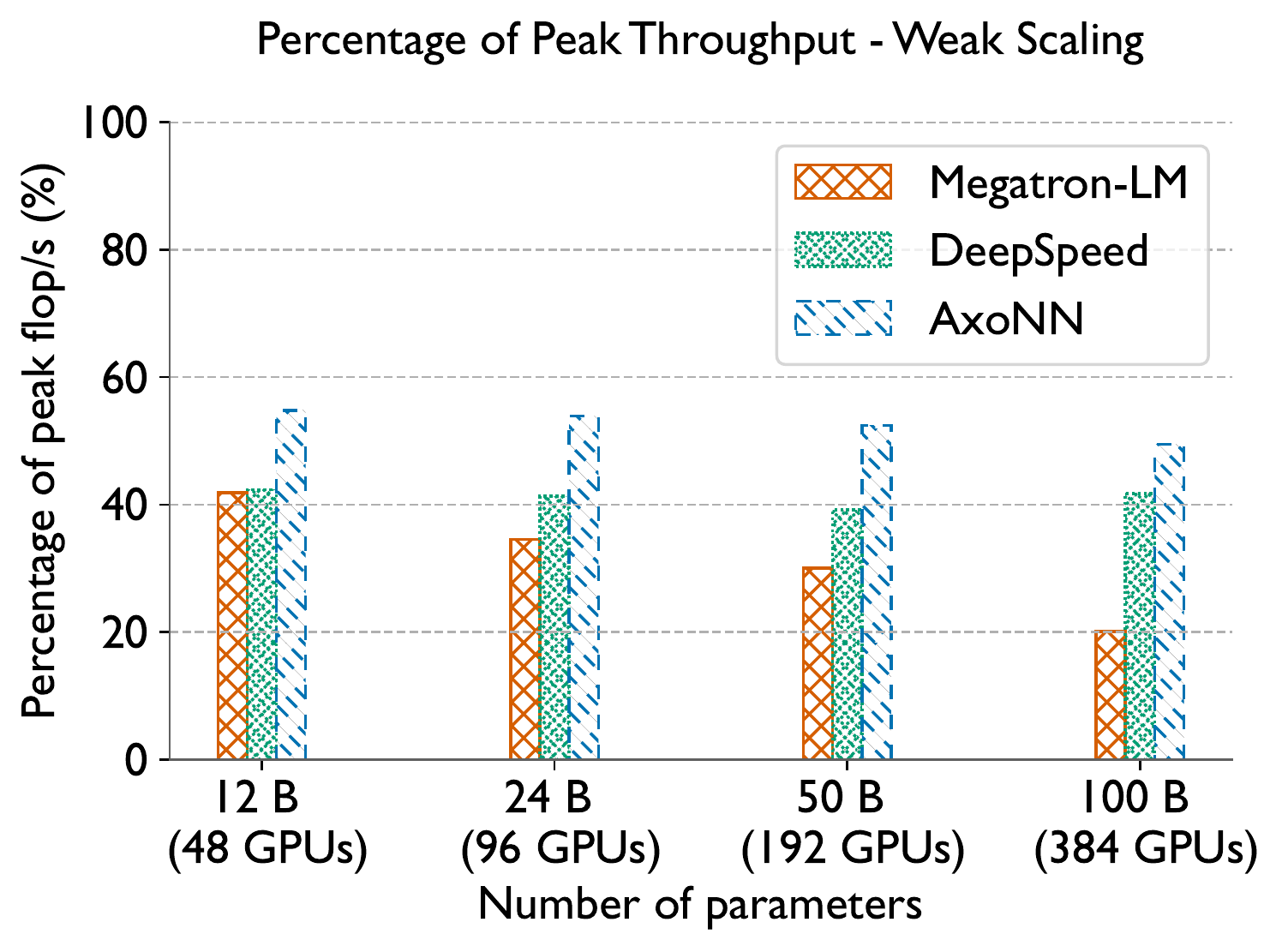}
    \caption{Plots comparing the weak scaling performance of AxoNN with other
frameworks: expected training times (left) and \% of peak GPU
throughput (right).}
    \label{fig:weakscaling}
\vspace{-0.1in}
\end{figure*}

\begin{figure}[h]
    \centering
      \includegraphics[width=\columnwidth]{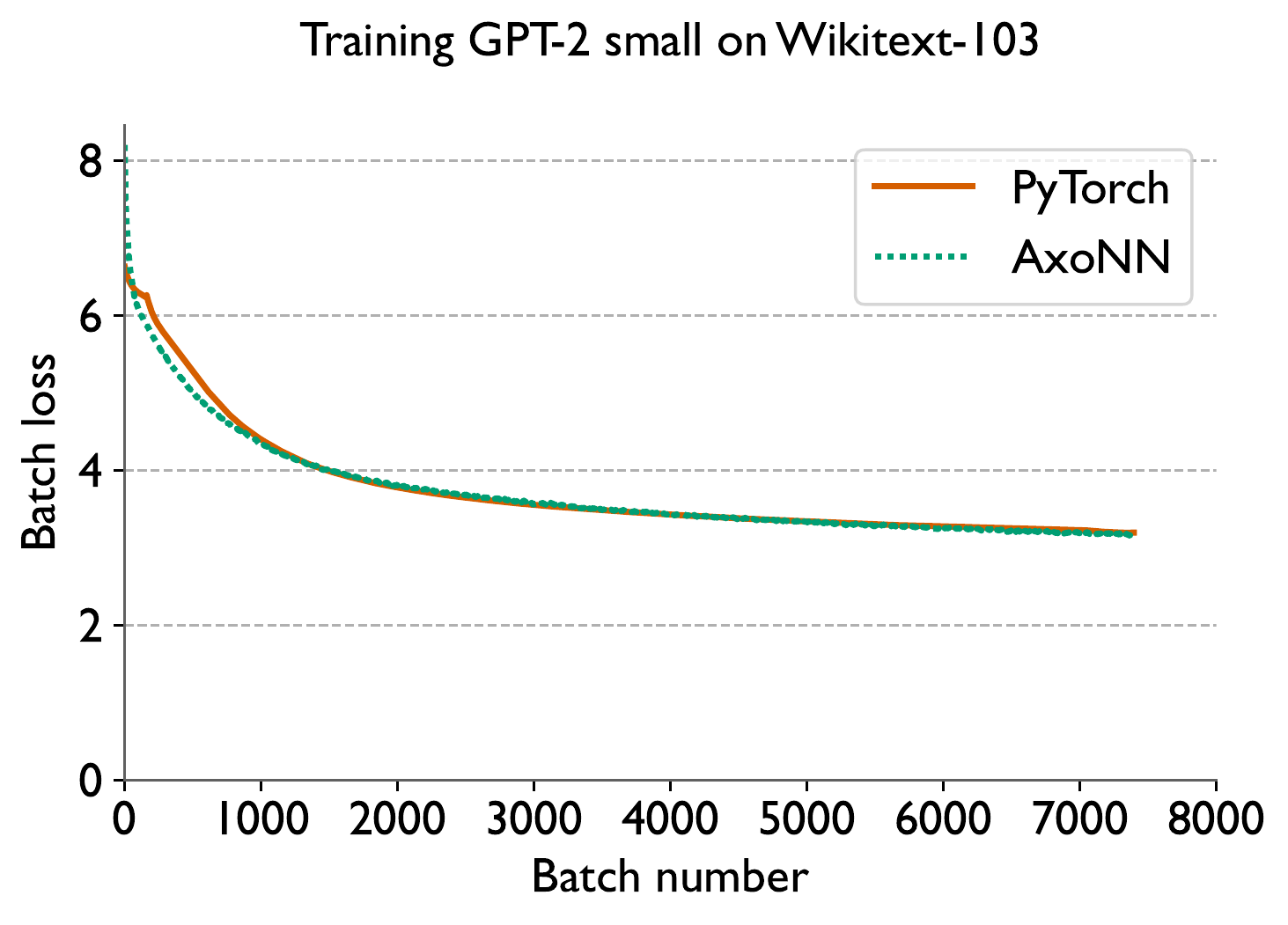}
    \caption{Loss curves for training GPT-2 small on the wikitext-103 dataset. We run \axonn on 12 GPUs ($G_\mathit{inter}=2$) and PyTorch on a single GPU.}
    \label{fig:stat-eff}
\vspace{-0.05in}
\end{figure} 

We now present the results of the experiments outlined in the previous section. 

\subsection{Training validation}

It is critical to ensure that parallelizing the training process does not
adversely impact its convergence. Diverging training loss curves are a sign of
undetected bugs in the implementation or statistical inefficiency of the
parallel algorithm. To validate the accuracy of our parallel implementation, we
train the 110 million parameter GPT-2 small to completion using PyTorch on a
single GPU and using \axonn on 12 GPUs with $G_\mathit{inter}=2$.
Figure~\ref{fig:stat-eff} shows the training loss for PyTorch, and \axonn and
we can see that the loss curves are identical. This validates our \axonn
implementation.

\subsection{Weak scaling performance}

To be fair to each framework, we tune various hyperparameters for each
framework on each GPU count and use the best values for reporting performance
results.  Table~\ref{tab:weakscale-hyperparams} lists the optimal
hyperparameters we obtain in our tuning experiments for each framework in the
weak scaling experiment. Across all model sizes, \axonn uses four to eight
times the number of GPUs for data parallelism as compared to Megatron-LM.  This
number is identical for \axonn and DeepSpeed for the 12 billion and 24 billion
parameter models but for the larger 50 and 100 billion parameter models, \axonn
uses twice as many GPUs for data parallelism as DeepSpeed. Since data
parallelism is embarrassingly parallel, this ends up substantially improving
\axonn's performance.

\begin{table}[h]
\centering
    \caption{\label{tab:weakscale-hyperparams}Optimal hyperparameter values obtained from tuning experiments for
the weak scaling studies.}
    \begin{tabular}{rlcrrc} \toprule
    \begin{tabular}[c]{@{}l@{}}No. of\\Params.\\(billions)\end{tabular} & Framework & \begin{tabular}[c]{@{}l@{}}Micro \\ Batch \\ Size\end{tabular} & $G_\mathit{intra}$ & $G_\mathit{inter}$ & $G_\mathit{data}$ \\ \midrule
    \multirow{3}{*}{12}  & \axonn    & 8                                                              & -           & 6           & 8          \\
                         & DeepSpeed & 2                                                              & 3           & 2           & 8          \\
                         & Megatron-LM  & 8                                                              & 3           & 16          & 1          \\ \midrule
    \multirow{3}{*}{24}  & \axonn    & 4                                                              & -           & 12          & 8          \\
                         & DeepSpeed & 2                                                              & 3           & 4           & 8          \\
                         & Megatron-LM  & 1                                                              & 3           & 16          & 2          \\ \midrule
    \multirow{3}{*}{50}  & \axonn    & 4                                                              & -           & 24          & 8          \\
                         & DeepSpeed & 1                                                              & 3           & 16          & 4          \\
                         & Megatron-LM  & 8                                                              & 6           & 32          & 1          \\ \midrule
    \multirow{3}{*}{100} & \axonn    & 2                                                              & -           & 48          & 8          \\
                         & DeepSpeed & 1                                                              & 3           & 32          & 4          \\
                         & Megatron-LM  & 4                                                              & 12          & 32          & 1          \\ \bottomrule
    \end{tabular}
\end{table}

Figure~\ref{fig:weakscaling} (left) presents a performance comparison of the
three frameworks in the weak scaling experiment.  When compared with the next
best framework - DeepSpeed, \axonn decreases the estimated training time by
over a month for the 12, 24 and 50 billion parameter models and 22 days for the
100 billion parameter model. For the 100 billion parameter model, \axonn is
faster than DeepSpeed by 1.18$\times$ and Megatron-LM by 2.46$\times$! This is
significant for deep learning research as it allows us to train larger models
faster.  Even at identical values of $G_\mathit{data}$ for the 12 and 24
billion parameter models, \axonn surpasses DeepSpeed because of our
asynchronous, message-driven implementation of inter-layer parallelism. These
results suggest that \axonn could scale to training trillion parameter neural
networks on thousands of GPUs in the future. \axonn also delivers an impressive
49-54\% of peak half precision throughput on Summit GPUs, outperforming
DeepSpeed (39-42\%) and Megatron-LM (21-41\%) (see
Figure~\ref{fig:weakscaling}, right).

\subsection{Strong scaling performance}

\begin{figure}[h] \centering
\includegraphics[width=\columnwidth]{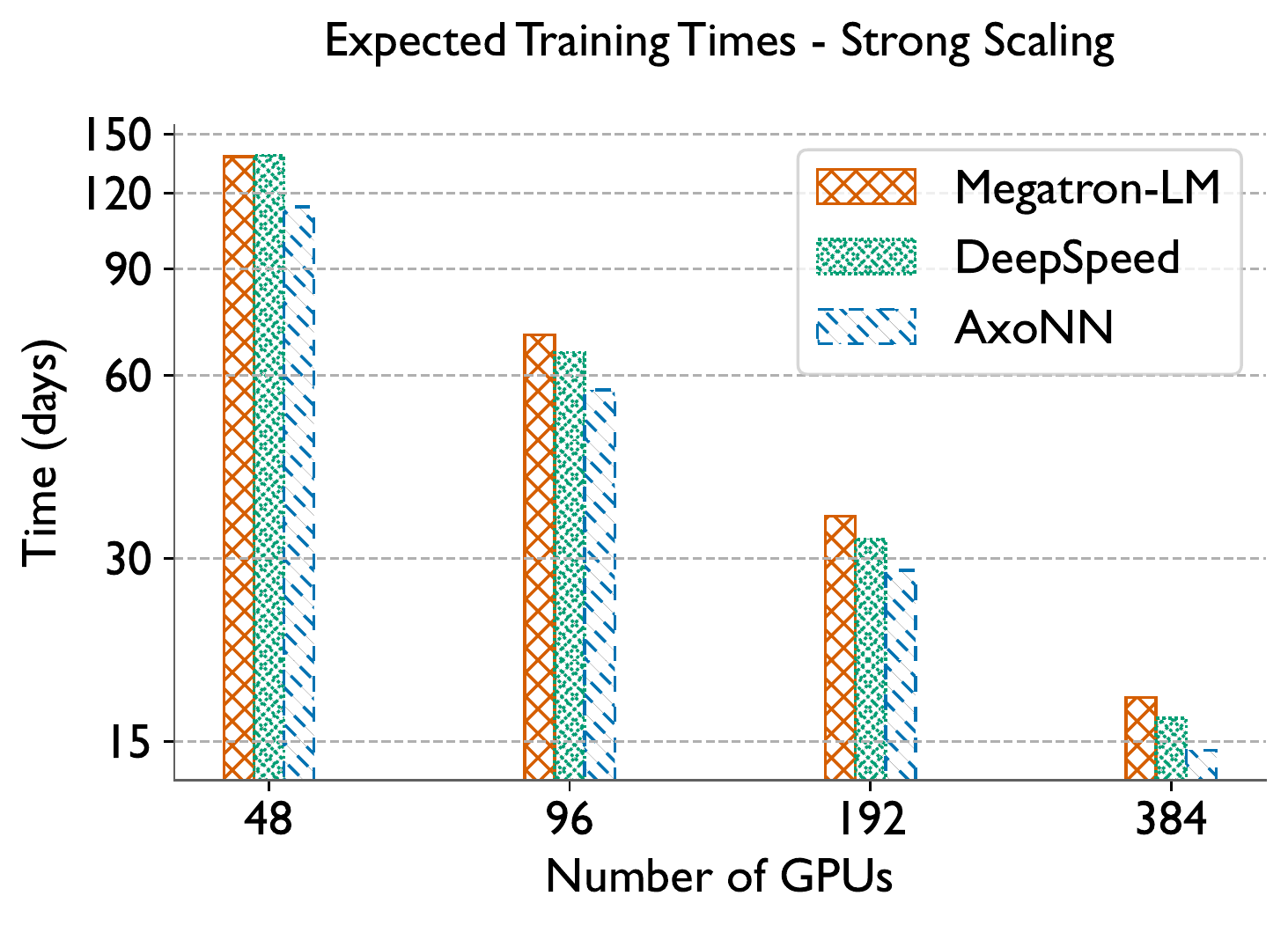}
    \caption{Plots comparing the strong scaling performance of AxoNN with other frameworks: training times for the 12 billion parameter transformer model.}
    \label{fig:strongscaling_expt}
\end{figure}

As with weak scaling, we first tuned hyperparameters for each framework for
strong scaling experiments. For these experiments we see all the three
frameworks using the same values of $G_\mathit{inter}$, $G_\mathit{intra}$ (not
applicable for \axonn) as Table~\ref{tab:weakscale-hyperparams} and scale the
value of $G_{data}$ with increasing GPU counts. This is a testament to data
parallelism's near perfect scaling behavior due to its embarrassingly parallel
nature. Figure~\ref{fig:strongscaling_expt} compares the strong scaling
performance of the three frameworks. Once again, \axonn again outperforms both
DeepSpeed and Megatron-LM by 11.47 and 18.14\% respectively on 384 GPUs.

\section{Related work}
Due to it's simplicity and embarrassingly parallel nature, data parallelism has
been the most commonly adopted algorithm in parallel deep learning research.
Initial frameworks gravitated towards asynchronous data parallelism with
parameter servers~\cite{projectadam,distbelief}. Chen et al however established
that asynchronous data parallelism does not work on a large number of
GPUs~\cite{chen2016revisiting}. The ensuing discrepancy between model weights
on each GPU ends up hurting the rate of convergence.  Subsequently, modern
implementations of data parallelism are synchronous and do not employ central
parameter
servers~\cite{pytorchdist-vldb,sergeev2018horovod,megatronlm,megatronlm-2,sc2020zero,zero_infinity}.
These frameworks average gradients using all-reduce communication primitives in
a bulk synchronous fashion after the backward pass of a batch is completed.
With advances in interconnect technology and communication
libraries~\cite{nccl}, the cost of synchronous all-reduce communication has
drastically reduced, making data parallelism the most effective algorithm for
scaling neural network training on 100s of GPUs.

Data parallelism needs to be combined with one or both of intra-layer and
inter-layer parallelism when the memory requirements of a neural network exceed
the DRAM capacity of a GPU. The exponentially increasing parameter sizes of
modern neural networks~\cite{gpt-2,gpt-3} have made it absolutely critical to
develop efficient algorithms for intra-layer and inter-layer parallelism.
While a number of frameworks have been proposed for intra-layer
parallelism~\cite{mesh_tf,megatronlm}, frequent collective communication calls
after the computation of each layer prevents them from scaling beyond a small
number of GPUs connected by NVLink.  Algorithms for inter-layer parallelism
fall into two categories based on the type of pipelining they implement: namely
pipelining with flushing~\cite{huang2019gpipe_nips,megatronlm-2,zero_3D,rannc}
or pipelining without flushing~\cite{narayanan2019pipedream,pipemare}. Under
the former approach, worker GPUs update their weights only after all of the
microbatches of a batch have been flushed out of the pipeline. While this
maintains strict optimizer semantics, constant flushing leads to inefficient
hardware utilization. This problem is greatly exacerbated at higher GPU counts.
Pipelining without flushing was proposed as a remedy for this problem. In this
approach a constant number of microbatches are always present in the pipeline.
Each GPU updates their weights asynchronously after completing the backward
pass of a microbatch. While this leads to increased hardware utilization, the
departure from exact optimizer semantics ends up hurting model convergence
severely. Again, greater the GPU count the more severe this problem is.
Subsequently, modern parallel deep learning frameworks have adopted pipelining
with flushing for realizing inter-layer parallelism. 

To counteract the extreme memory requirements of modern neural networks,
Rajbhandari et al.~have augmented data parallelism with a number of memory
optimizations (ZeRO~\cite{sc2020zero}, ZeRO-Offload~\cite{zero_offload},
ZeRO-Infinity~\cite{zero_infinity}). ZeRO distributes optimizer states,
parameters and gradients across data-parallel GPUs. ZeRO-Offload and
ZeRO-Infinity are targeted towards extremely memory scarce environments. To
reduce GPU memory utilization they offload data to the CPU or NVMe.

\section{Conclusion}
In this work, we presented a new highly scalable parallel framework for deep
learning, called \axonn. We have demonstrated that \axonn utilizes available
hardware resources efficiently by exploiting asynchrony and message-driven
scheduling. We augmented \axonn with a novel memory optimization algorithm that
not only provided a four-fold savings in GPU memory utilization, but also
boosted performance by over 13\%. In both strong and weak scaling experiments,
\axonn outperformed the state-of-the-art for training large multi-billion
parameter transformer models.  We believe that \axonn will allow deep learning
researchers to save valuable resources and time in their training runs. Our
results give us hope that we can use \axonn to train transformer models with
more than one trillion parameters on thousands of GPUs. We plan to open-source
the weights of these trained models for the benefit of the research community.

\section*{Acknowledgement}
This work was supported by funding provided by the University of Maryland
College Park Foundation.  This research used resources of the Oak Ridge
Leadership Computing Facility at the Oak Ridge National Laboratory, which is
supported by the Office of Science of the U.S.~Department of Energy under
Contract No.~DE-AC05-00OR22725.

\bibliographystyle{IEEEtran}
\bibliography{./bib/cite,./bib/pssg}

\end{document}